%% file: acl_latex.tex
\pdfoutput=1
\documentclass[11pt]{article}

\usepackage{acl}
\usepackage{booktabs}
\usepackage{times}
\usepackage{latexsym}
\usepackage{amsfonts}       % blackboard math symbols
\usepackage{nicefrac}       % compact symbols for 1/2, etc.
\usepackage{microtype}      % microtypography
\usepackage{xcolor}    
\usepackage[listings]{tcolorbox}
\tcbuselibrary{listings,theorems,breakable}

\newtcolorbox{mybox}{colback=white!5!white,colframe=black!75!black, left=.05in, right=.05in}

\newtcbtheorem[number within=section]{exmp}{Prompts}%
{breakable,colback=white!5!white,colframe=black!95!,fonttitle=\bfseries, left=.02in, right=.02in,bottom=.02in, top=.02in}{exmp}

\newtcbtheorem[number within=section]{case}{Examples}%
{breakable,colback=white!5!white,colframe=black!95!,fonttitle=\bfseries, left=.02in, right=.02in,bottom=.02in, top=.02in}{case}

\usepackage{float} 
\usepackage[export]{adjustbox}
\usepackage{dashrule}
\usepackage{multirow}
\usepackage{multicol}
\usepackage{graphicx}
\usepackage{amsmath}
\usepackage{amssymb}
\usepackage{amsfonts}

\usepackage{colortbl}
\definecolor{light}{gray}{.85}
\definecolor{title}{gray}{.30}

\usepackage[T1]{fontenc}
\usepackage[utf8]{inputenc}

\usepackage{microtype}

\usepackage{inconsolata}

\usepackage{graphicx}

\title{ControlMath: Controllable Data Generation Promotes Math Generalist Models}

\author{
 Nuo Chen$^\spadesuit$
\quad
Ning Wu$^\clubsuit$
\quad 
{\bf Jianhui Chang$^{\diamondsuit}$} 
{\bf \quad Jia Li$^\spadesuit$}\\
\\
  $^\spadesuit$Hong Kong University of Science and Technology (Guangzhou)\\ Hong Kong University of Science and Technology\\
  $^\clubsuit$Microsoft \\
  $^{\diamondsuit}$China Telecom Cloud Computing Research Institute\\
    \texttt{nchen022@connect.ust.hk},
    \texttt{changjh1@chinatelecom.cn},
    \texttt{jialee@ust.hk}\\}
\begin{document}
\maketitle
\begin{abstract}
Utilizing large language models (LLMs) for data augmentation has yielded encouraging results in mathematical reasoning. However, these approaches face constraints in problem diversity,  potentially restricting them to in-domain/distribution data generation. 
To this end, we propose \textbf{ControlMath}, an iterative method involving an equation-generator module and two LLM-based agents. The module creates diverse equations, which the Problem-Crafter agent then transforms into math word problems. The Reverse-Agent filters and selects high-quality data, adhering to the "less is more" principle, achieving better results with fewer data points. This approach enables the generation of diverse math problems, not limited to specific domains or distributions. 
As a result, we collect ControlMathQA, which involves 190k math word problems. Extensive results prove that combining our dataset with in-domain datasets like GSM8K can help improve the model's mathematical ability to generalize, leading to improved performances  both within and beyond specific domains.

\end{abstract}

\input{Section/1Introduction}

% \section{Introduction}
\input{Section/2RelatedWorks}

\input{Section/3Method}

\input{Section/4Experiments}

\input{Section/5Conclusion}
\section*{Limitation}
In this paper, we focus on using large language models to generate training samples for mathematical reasoning, Some issues remain to be explored:

In practice, we observe that \texttt{gpt-4-turbo} may generate unrelated questions from the given equations, especially for complex ones involving 7 or 8 steps. We found that the error rate for such cases can be as high as 25-35\%, leading to significant cost inefficiencies. A more efficient solution needs to be explored to address this issue.

Another limitation is that we have not yet conducted experiments on larger LLMs like LLaMA 2-30B or 70B models , primarily due to training resource constraints.

Additionally, due to the specificity of our method, we can generate large amounts of data tailored to specific datasets, potentially causing the model to overfit within that domain and thus achieve better performance. This could lead to unfair advantages in leaderboard rankings, though it ultimately depends on the practitioners' professional ethics.

\bibliography{custom}
% \clearpage
\appendix

\section{Prompts}

\label{sec:appendix}

In this section, we show the prompts for Problem-Rewriter agents and training.

\input{tables/prompts}

% \section{Cases}

\input{tables/cases}
\end{document}

%% file: Section/1Introduction.tex
\section{Introduction}

% Large Language Models (LLMs) \cite{DBLP:journals/corr/abs-2005-14165, hu2021lora,  openai2023gpt4} have shown exceptional performance across a wide range of tasks, such as open-ended dialogue generation \cite{zeng2022glm}, code generation \cite{gao2023pal}. However, 

Currently, mathematical reasoning \cite{DBLP:journals/corr/abs-2110-14168, chen2023breaking, zhou2022least,weng2022large} is regarded as the  one of the most challenging areas for current Large Language Models (LLMs). Typically, prompting-based approaches \cite{wei2022chain, DBLP:journals/tmlr/WeiTBRZBYBZMCHVLDF22, wang2022self} are common ways to improve the mathematical abilities of closed-source LLMs. These methods design different prompts for these  LLMs to solve multi-step and complicated math problems, setting a high benchmark and demonstrating the potential of LLMs in tackling sophisticated mathematical problems.

\begin{figure}
    \centering
    \includegraphics[width=1\linewidth]{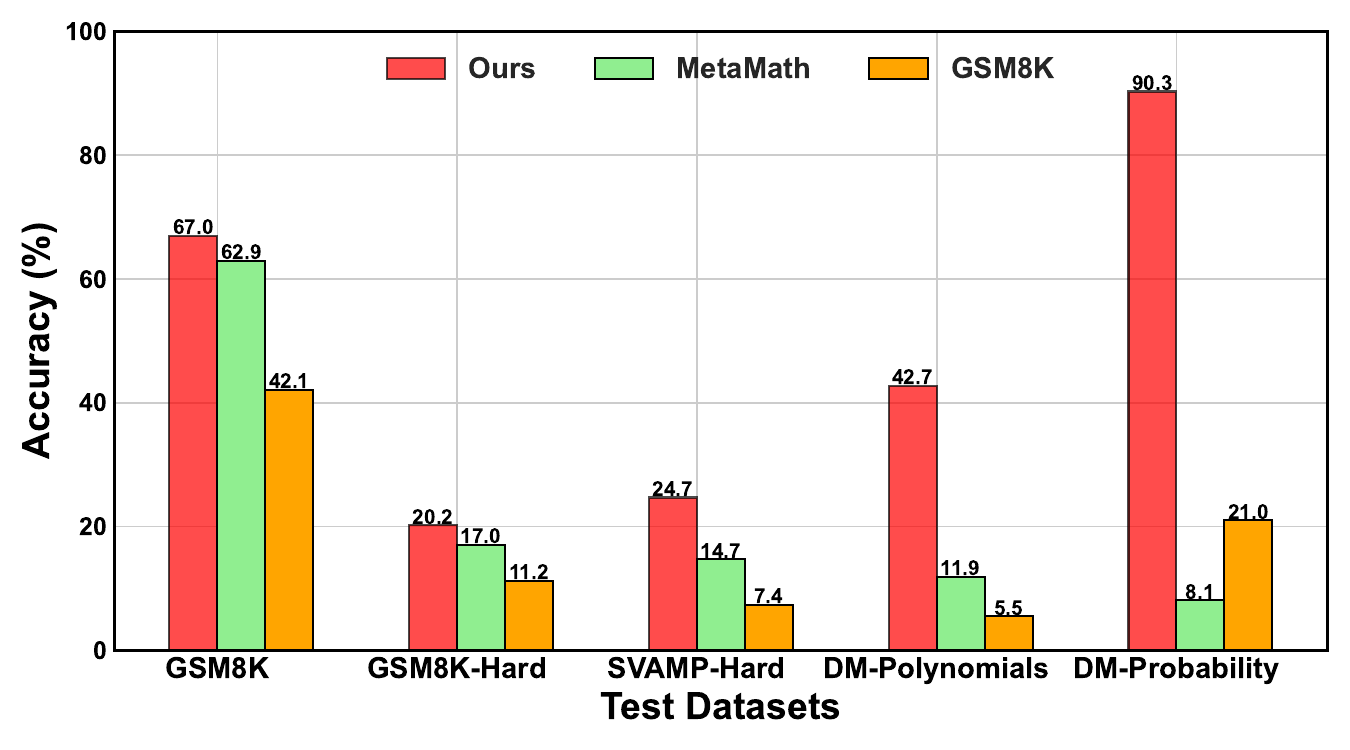}
    \caption{We train LLaMA 2-7B model \cite{llama2} with different training corpus and present results in out-of-domain/distribution datasets: GSM8K-Hard \cite{gao2023pal}, SVAMP-Hard \cite{chen2023good}, DM-Polynomials and Probability subdatasets \cite{saxton2018analysing}.}
    \label{fig:motivation}
    \vspace{-15pt}
\end{figure}

Recently, the focus has shifted towards improving the capabilities of smaller, open-source LLMs through instruction-tuning \cite{DBLP:journals/corr/abs-2309-12284, 10.1145/3637528.3672010, DBLP:conf/nips/Ouyang0JAWMZASR22,peng2023instruction,zhang2023instruction,longpre2023flan,toshniwal2024openmathinstruct}. A significant advancement in this area is the use of data augmentation, using frontier LLMs like ChatGPT or GPT-4  to increase the size of available datasets to improve model performance \cite{yuan2023scaling, DBLP:journals/corr/abs-2309-12284, li2023query, tang2024mathscale}. 
% Notably, this augmentation often leverages frontier LLMs like ChatGPT or GPT-4 to generate ne. 
For example, MetaMath \cite{DBLP:journals/corr/abs-2309-12284} and MuggleMath \cite{li2023query} employ  different augmented methods, such as question rephrasing, on the GSM8K \cite{DBLP:journals/corr/abs-2110-14168} or MATH \cite{hendrycksmath2021} datasets to generate new problems.  
However, these methods depend heavily on seed questions from the training datasets, resulting in new samples that closely resemble the original ones. This lack of diversity can result in models overfitting to specific domains or data distributions, as the new problems maintain similar topics, reasoning steps, and numerical operations as the original ones. Thus, the models are effective within certain domains/distributions but less capable outside them \cite{li2023query}.

Figure \ref{fig:motivation} showcases the effects of in-domain data augmentation using MetaMath on the GSM8K dataset. MetaMath achieves a significant performance boost on GSM8K, increasing accuracy by over 20\%. However, when tested on out-of-domain datasets like SVAMP-Hard, DM-Polynomials, and DM-Probability, the model's   performance drops sharply, sometimes even below the baseline trained only on GSM8K. Notably, on the GSM8K-Hard dataset, which only varies the distributions of the numbers, there is also a substantial decline in accuracy. This highlights that while MetaMath improves in-domain performance, it struggles to generalize to more diverse or challenging datasets.

Moreover, another critical oversight in previous approaches is the inclusion of low-quality or redundant samples in augmented datasets. Although methods like MetaMath have significantly expanded original datasets from 14k to 400k, concerns remain about the quality and utility of all generated samples. Some samples may be of low quality or redundant, failing to contribute meaningfully to the model’s learning process and increasing useless training costs \cite{DBLP:journals/corr/abs-2307-08701}. This highlights the need for an effective strategy to filter out ineffective training samples, a challenge still unaddressed in mathematical reasoning.

To address the above challenges, we propose a  simple and iterative approach for controllable mathematical data  generation, termed \textbf{ControlMath}. The core concept is \textit{from equation to math word problem} and the 
essence of control lies in two primary perspectives: (1) Controllable equation generation: By specifying requirements such as reasoning steps, designed operators, and numerical ranges, we generate corresponding equations through an equation generation module. The frontier  LLM-based Problem-Crafter Agent then creates mathematical word problems based on these equations, ensuring control over the diversity and distribution of the final set of problems. (2) Adaptively efficient data selection: This is achieved through a Problem Rewriter-Agent to ensure the effectiveness of the training samples. This agent rewrites the generated mathematical problems to create variations. Both the original and rewritten problems are fed to the LLM; if the LLM solves both correctly, it indicates that this generated problem does not contribute to further learning and can be discarded. This approach mimics human cognitive learning, where redundant information does not enhance learning efficiency. Filtering out redundant samples allows us to focus on the most beneficial data, enhancing overall training efficiency. The overview of our approach is presented in Figure \ref{fig:framework}. 

By controlling the generation and selection of training samples, our method embodies the principle of "less is more", focusing on quality over quantity to enable stronger models with smaller datasets. To prove its effectiveness, we curate a dataset-ControlMathQA-comprising approximately 190k training samples. By seamlessly integrating this dataset with established GSM8K and MetaMath datasets, we embark on training LLMs. Our empirical results prove the resulting generalist LLMs clearly show significant performance improvements in both in-domain and out-of-domain datasets, as shown in Figure \ref{fig:motivation}.  In general, our contributions are summarized as follows:

\begin{itemize}
    \item We propose a new data generation strategy, named ControlMath, for mathematical reasoning, which consists of controllable equation generation and an adaptively efficient data selection process.
    \item We collect ControlMathQA, a new training corpus that involves about 190k samples to help build math generalist models.
    \item We further investigate different variants of ControlMath to show its scaleability and generalization.
\end{itemize}

% This shows that our proposed method is naturally effective and flexible.

%% file: Section/2RelatedWorks.tex
\section{Related Works}

\begin{figure*}
\vspace{-15pt}
    \centering
    \vspace{-5pt}
    \includegraphics[width=1\linewidth]{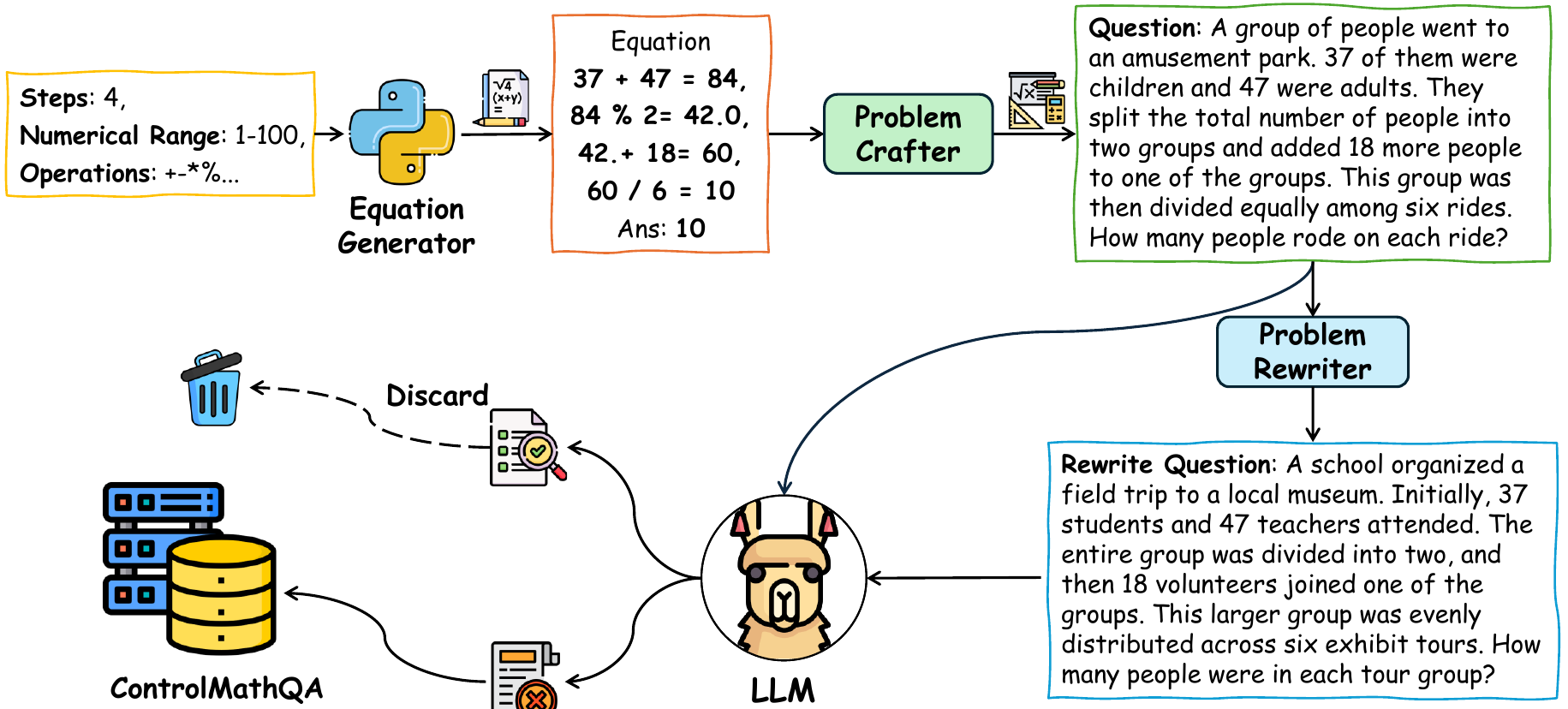}
    \caption{The overview of our ControlMath. Here, we present the example of generating multi-calculation math word problems.}
    \label{fig:framework}
    \vspace{-15pt}
\end{figure*}

\paragraph{Mathematical Reasoning with LLMs.} In recent times, large language models (LLMs) \cite{DBLP:journals/corr/abs-2005-14165, hu2021lora,  openai2023gpt4, touvron2023llama,huang2024large, chen-etal-2023-large} have demonstrated remarkable abilities in handling complex mathematical reasoning tasks (MR) \cite{scao2022bloom, DBLP:journals/corr/abs-2110-14168, zhou2022least, weng2022large, chen2023teaching,imani2023mathprompter}. Two primary approaches are used: (1) Prompting-based methods \cite{DBLP:conf/acl/ImaniD023, wang2022self, gao2023pal,shi2022language,chen2023teaching}, which employ diverse prompts to aid LLMs in solving mathematical problems. A notable example is Chain-of-Thought (CoT) \cite{wei2022chain}, which guides LLMs through step-by-step prompting. (2) Finetuning-based  approaches \cite{li2023making, yuan2023scaling}, which unlock the potential of open-source LLMs for mathematical reasoning through instruction-tuning, relying on effective downstream training data.

\paragraph{Data Augmentation with LLMs.} In the era of LLMs, data augmentation is regarded as a useful approach to improve smaller LLMs' performances in down-stream tasks. This approach often involves distilling outputs of stronger LLMs to generate the SFT datasets \cite{luo2023wizardmath, mitra2024orca, yin2024mumathcode,yue2024mammoth,meng2022generating}. When applying data augmentation into math reasoning, there are two variants: query augmentation and response augmentation. Typical works towards query augmentation include MuggleMath \cite{li2023query}, MetaMath \cite{DBLP:journals/corr/abs-2309-12284} and MathScale \cite{tang2024mathscale}, which use different strategies to obtain the new questions.
% MuggleMath \cite{li2023query}, MetaMath and MathScale \cite{tang2024mathscale} mainly focus on query augmentation, which use different strategies to obtain the new questions. 
% For instance,  MuggleMath creates new questions by replacing concepts and numbers in the GSM8K dataset.
Meanwhile, RFT \cite{yuan2023scaling} and xRFT \cite{chen2023breaking} concentrate on response augmentation, which utilize rejection sampling to generate diverse reasoning paths for the same question. However, most of these approaches rely on seed questions in training sets, which pose challenges in generating diverse domain or distribution data.

%% file: Section/3Method.tex
\section{ControlMath}

The primary motivation behind ControlMath is to address the challenges of overfitting and inefficiency in training LLMs for mathematical reasoning tasks. 
% Current data augmentation methods often result in training data that lack diversity, leading to models that perform well within specific domains/distriburions but struggle with broader applications.
To achieve this, ControlMath employs a three-step approach involving two specialized agents and one Python function module. 
In this section, we present the details of ControlMath.

\subsection{Controllable Equation Generation}
\label{subsec:equation}
Our first step is the controllable generation of mathematical equations, driven by the need to ensure diversity in the following generated math word problems. Specifically, we employ an equation generation module to allow precise control over the characteristics of the equations, addressing the issue of homogeneous training samples that limit the model's ability to generalize.

In our implementation, the equation generation module is a Python function designed to produce equations based on specified parameters. We focus on generating three main types of equations: 1) \textbf{Multi-step Calculation}: These include basic arithmetic operations (addition, subtraction, multiplication, division), as well as more complex functions like trigonometric functions, roots, and exponents. For generating such equations, we specify the input parameters, including \textit{steps}, \textit{operators} and \textit{numerical ranges}, to create diverse equations; 2) \textbf{Polynomials}: These equations involve terms with variables raised to various powers, often requiring multiple steps to simplify or solve. By generating polynomials of varying \textit{degree} of the polynomial and the \textit{coefficients}, we ensure that the model encounters a wide range of polynomial-related problems; 3) \textbf{Probability}: These equations involve calculating probabilities, which often require understanding combinations, permutations, and probability rules. We follow \cite{hendrycksmath2021} and set predefined topics: \textit{a set of letters} or \textit{sequence} as templates. These templates serve as the basis for generating diverse probability problems, ensuring the LLMs develop a strong foundation in statistical reasoning.

% Each type equation involves

By customizing different inputs, we generate a wide variety of formulas that cover a broad spectrum of mathematical concepts and operations. This controlled generation ensures the foundation for creating diverse mathematical word problems.

\input{tables/crafter}

% Unlike previous methods that modify existing math word problems, we directly generate diverse equations by controlling the \textit{steps}, \textit{operators}, and \textit{numbers} involved in these equations. 

\input{tables/example}

\paragraph{Problem-Crafter Agent.} The second step involves translating the abstract equations into math word problems.  A frointer LLM-based Problem-Crafter  agent is used for this purpose, which generates \textit{contextually rich problems} and their \textit{reference solutions} from the generated equations. It ensures that problems vary in context and application while maintaining a strong connection to the underlying equations.
We present the template for prompts in Table \ref{table:crafter}. To guide the agent in question formulation, we include several examples in the prompts, which are from manual interactions between Web GPT-4.

\subsection{Efficient Data Selection: More data is not always helpful}   
\label{subsec: question}

\paragraph{Problem-Rewriter Agent.} The third step focuses on selecting effective training data and eliminating redundant or ineffective samples. The motivation for this step is to enhance training efficiency by ensuring that only useful data is retained, thereby improving model performance without overfitting. The Problem-Rewriter agent is crucial for this step. After generating an initial problem by the Problem-Crafter, this agent rewrites the problems to create variations. Note that the rewriter agent is constrained to maintain the original  numerical values of the problems, changing only the phrasing or topics, as shown in Appendix \ref{sec:appendix}.

\paragraph{Efficient Data Selection.} Then, both the original and rewritten math word problems are presented to the smaller open-sourced LLM; if the LLM can solve both correctly, it means such  problem does not contribute to further learning and can be discarded  because model may
have mastered this level of problems. However, if the model answers  incorrectly to one of them, it indicates a genuine gap in understanding, making these problems valuable for further training.  Such problem will be collected in our \textbf{ControlMathQA}. 
This process mimics human cognitive learning, where exposure to redundant information does not enhance understanding. By focusing on non-redundant, effective samples, we improve the overall quality and efficiency of the training data. To reduce variance, the temperature is set to 0 when sampling LLM's outputs. 
% Note that, we only collect original problems as we find that including rewritten problems contributes  limited  improvements (See results in Appendix, Figure \ref{fig:rewritten}).

\paragraph{Validation. }
In practice, we observe that sometimes the new problems generated by the Problem-Crafter Agent might not correctly match the corresponding formulas or the generated reference solutions might be inaccurate. To address these issues, we introduce a validation step using GPT-4 (\texttt{gpt-4-turbo}).  For validation, we input the formula, the generated problem, and the corresponding solution into GPT-4. If GPT-4 identifies any mismatches—either the problem does not align with the formula, or the solution does not correctly solve the problem—we discard the problematic entry. This validation step ensures that only accurate and relevant problems are included in our dataset, further enhancing the quality of the training data. The prompts are presented in Appendix \ref{sec:appendix}.

\paragraph{Iterative Process.} In theory, our approach can be iteratively applied indefinitely to continually generate needed corpus. However, due to resource constraints, we limit  experimentation to \textbf{three iterations}. In the first iteration, we use Mistral-7B which has not undergone any fine-tuning. For the second and third iterations, we fine-tune the model using the data generated from the previous iterations. This iterative process allows us to progressively enhance the model's performance by continually refining the training data and focusing on the model's weaknesses identified in each round.

\paragraph{Why we need to rewrite the problem?} The purpose of rewriting the problem is to ensure the robustness of the model's understanding. Several works \cite{chen2024premise, DBLP:journals/corr/abs-2309-12284} have proved that simply rephrasing or  changing the order of the question could confuse LLMs. Thus, simply testing the model on the original problem might not fully capture its comprehension, as the model could be overfitting to specific phrasings or patterns in the training data. By presenting both the original and the rewritten problem, we can more accurately assess whether the model truly understands the underlying concepts. We have also complie a confusion matrix showing the proportions of cases where the model answers both questions correctly, both incorrectly, or one correctly and one incorrectly, detailed in the Appendix, Table \ref{pilot}.

% \paragraph{Takeaways.} From a more targeted application perspective, ControlMath's adaptability allows it to focus on specific mathematical datasets or tasks by selectively enhancing data in areas where the model shows deficiencies. For instance, if a model underperforms on certain types of problems within a dataset, ControlMath can generate additional targeted problems to address these weaknesses. This targeted data augmentation ensures the superior  performances on specific task. More experiments and discussions are presented in Section \ref{sec:discuss}.

\paragraph{Correlation with AdaBoost.} Our approach is theoretically similar to AdaBoost \cite{freund1999short}, a machine learning algorithm that iteratively trains weak classifiers on the hardest-to-classify samples, reweighting data to focus on errors. Similarly, ControlMath focuses on generating and selecting the most challenging problems for the LLM, effectively "reweighting" the training data towards the areas where the model struggles the most. By collecting only the problems the model answers incorrectly, we ensure that subsequent training rounds address these weaknesses, analogous to how AdaBoost emphasizes misclassified samples to improve overall model performance. This similarity to AdaBoost gives ControlMath a  theoretical base, showing that it has the potential to make LLMs more reliable  in general mathematical reasoning.
% By systematically addressing the model's weaknesses through controlled data generation and effective problem selection, ControlMath enhances learning efficiency.

\subsection{Statistics}

In each iteration, we generate 160k multi-step calculation formulas, 10k polynomials and 10k probability problems (See details in following Section \ref{sub:data_generation}). After data selection and GPT-4 validation process, we collect 110k math word problems in total.
Inspired by \cite{yuan2023scaling}, response augmentation also contributes to the benefit of math reasoning LLMs, we also include this type augmentation in our ContronlMathQA, where we utilize \texttt{GPT-3.5-Turbo-0613} to generate answers for each question three times and remain the correct ones. As a result, we collect about \textbf{190k} question-answer pairs, where 155k for multi-calculation, 15k for polynomials and 20k for probability problems. Note, we decide to keep only original problems as we find  including rewritten ones contributes  limited  improvements (Results in Appendix, Figure \ref{fig:rewritten}).

%% file: tables/crafter.tex
\begin{table}[!t]\footnotesize
\centering
\small

\begin{tabular}{|p{0.95\linewidth}|}
% \begin{tabular}{lc}
\hline
\\
You are a math expert.  Given an equation list that contains multi-step mathematical  formulas and its final answer.
Your task is to create a math word problem and its reference solution based on the provided series of mathematical equations and the final answer.
\\

\\
\{\texttt{few shot examples}\}

\\
Equation List: \{\texttt{Equation}\} \\
Final Answer: \{\texttt{Answer}\} \\
\\

You should keep in mind that: \\
(1) Your produced math problem should be diverse and avoid using words like ``mathematical'', ``calculation'', etc. \\
(2) You should present equations in LaTeX format. \\

\hline
\end{tabular}

% English translation is only given for reading.} 
\caption{Prompt Template for Problem-Crafter Agent. }
\label{table:crafter}
\vspace{-5mm}
\end{table}

% \begin{tcolorbox}[
% colback=white!10!white,
% colframe=black!75!black,
% title=Fine-tuning system prompt,
% breakable]
% Below is an instruction that describes a task. Write a response that appropriately completes the request.\#\#\# Instruction:
%  \textbf{**Query.**} \#\#\# Response:
% % \caption{Query augmentation prompt.}
% % \label{tab:query-aug prompt}
% \end{tcolorbox}

%% file: tables/example.tex
\begin{table*}[!t]\footnotesize
\centering
\small
\vspace{-15pt}

\begin{tabular}{l|p{0.7\linewidth}}
% \begin{tabular}{lc}
\toprule
\rowcolor[gray]{.75}
\textbf{Specifications} & \textbf{Steps}: 4, \textbf{Operators}: [sqrt(), \(\times\), \%, +, -], \textbf{Numerical Range}: 1-1000 \\
% \midrule
\rowcolor[gray]{.75}
\textbf{Generated Equation} &  4 - 1 = 3, 3 * 16 = 48, sqrt(16) = 4, 30 \% 6=5\\
\midrule

\multirow{5}{*}{\textbf{Math Word Problem}} &Tom sets sail on his ship at 1 PM, traveling at a speed of 10 miles per hour. He reaches his destination at 4 PM. After spending some time there, he decides to return. However, due to different weather conditions, he sails back at a slightly slower speed that is the square root of the original speed.
Using this information, can you determine how long it takes Tom to return to his starting point?\\
\midrule
\midrule
% Operators: [sqrt, \(\times\), \%, +, -] &-  \\
% Numerical Range: 1-100 & -
% \\
\rowcolor[gray]{.75}
\textbf{Specifications} & \textbf{Degree}: 2,  \textbf{Coefficients}: -100-100 \\
% \midrule
\rowcolor[gray]{.75}
\textbf{Generated Equation} &  -41*c - 16*c**2 + 18*c + 25*c\\
\midrule
% \rowcolor[gray]{.75}
\multirow{1}{*}{\textbf{Math Word Problem}} & Express -41*c - 16*c**2 + 18*c + 25*c in the form q*c**2 + p*c + u and give p.\\
% \midrule
% \midrule
% \textbf{Specifications} & \textbf{Degree}: 3,  \textbf{Coefficients}: -100-100 \\
% % \midrule
% \textbf{Generated Equation} &  10*x**3 + 8*x**2 + 7*x + 1\\
% \midrule
% \multirow{5}{*}{\textbf{Math Word Problem}} & Alex is a computer programmer who is developing a new type of video game that simulates space travel. To model the changing velocity of a spacecraft, Alex uses a polynomial to represent the velocity (in kilometers per second) as a function of time 
% t in seconds after launch. The polynomial is v(t) =  10*t**3 + 8*t**2 + 7*t + 1. If Alex wants to know the velocity after 2 seconds, what will it be according to the model?
% \\
\midrule
\midrule
\rowcolor[gray]{.75}
\textbf{Specifications} & \textbf{Templates}: A set of letters,  \textbf{Unique characters}: 6, \textbf{Without Replacement} \\
\midrule

\multirow{2}{*}{\textbf{Math Word Problem}} & What is probability of picking 1 k, 1 h, and 1 c when three letters picked without replacement from \{c: 1, y: 1, e: 1, n: 1, k: 1, h: 2\}?\\

% Express -41c - 
% \[16c^{2}\]
%  + 18c + 25c in the form \(qc^{2}\) + pc + u and give p.

\bottomrule
\end{tabular}

% English translation is only given for reading.} 
\caption{Several generated examples in our ControlMathQA, including input specifications, generated equation and math word problems. More cases could be seen in Appendix \ref{sec:cases}.}
\label{table:example}
\vspace{-5mm}
\end{table*}

%% file: Section/4Experiments.tex
\section{Experiments}
\input{tables/main_results}
In this section, we first present the details of our implementation, including data generation, training settings. Then we show the main results of different backbones with ControlMathQA.

\subsection{Implementation}
\label{sub:data_generation}

\subsubsection{Data Generation}  In  equation generation (Section \ref{subsec:equation}), we introduce several constraints to ensure the generated equations are both high in quality and diverse: (1) \textbf{Unique equations}:  To secure robust samples, it is essential to keep each generated equation distinct; (2) \textbf{Diverse Distributions}: \textit{a}) For \textit{multi-step calculation formulas}, we set the range for steps from 2 to 9, covering most multi-step reasoning scenarios. The numerical range is divided into four tiers: 1-100, 100-k, 1k-10k, and 10k-1million. During each iteration, these ranges are randomly combined  to produce 5,000 unique equation examples for each combing group, with operators including basic arithmetic (addition, subtraction, multiplication, division), square roots, and exponents. \textit{b}) For \textit{polynomials}, we set the degree range from 1 to 3 and coefficients from -100 to 100, producing polynomials of different complexities, ensuring the LLMs encounter a broad range of polynomial-related problems. Each iteration randomly samples 5,000 polynomial equations. \textit{c}) Following \citet{saxton2018analysing}, we consider two sampling settings: with or without replacement to generate \textit{probability problems}. The sample space templates consist of a set of characters and a sequential sequence, where the number of unique characters does not exceed 20. In each interaction, we randomly sample 5000 examples. Of note, this setting directly produces math word problems, eliminating the need for the problem-crafter agent, as an example in Table \ref{table:example}. We use \textit{math}, \textit{sympy}, \textit{mathematics\_dataset} libraries to generate equations with python.
% A generated example is: ``What is prob of picking 1 k, 1 h, and 1 c when three letters picked without replacement from \{c: 1, y: 1, e: 1, n: 1, k: 1, h: 2\}?''

% \textit{FGenerating probability problems} involves using established settings from mathematics datasets \cite{saxton2018analysing}, with sequences or sets of letters as sample space templates. We set the distinct elements in total sample space to 20. The creation of problems involves producing 5,000 samples for each condition by randomly sampling with and without replacement from the entire sample space. This setting directly produces math word problems, eliminating the need for the problem-crafter agent. A generated example is: ``What is prob of picking 1 k, 1 h, and 1 c when three letters picked without replacement from \{c: 1, y: 1, e: 1, n: 1, k: 1, h: 2\}?''

In Section \ref{subsec: question}, both Problem-Crafter and Rewritten Agents are based on \texttt{gpt-4-turbo}. Key implementation details include setting the temperature to 0.9 to ensure diverse problem generation, max tokens to 2,000 to accommodate complex problem formulations, and top-p to 1. 

\begin{figure*}
    \centering
    \includegraphics[width=1\linewidth]{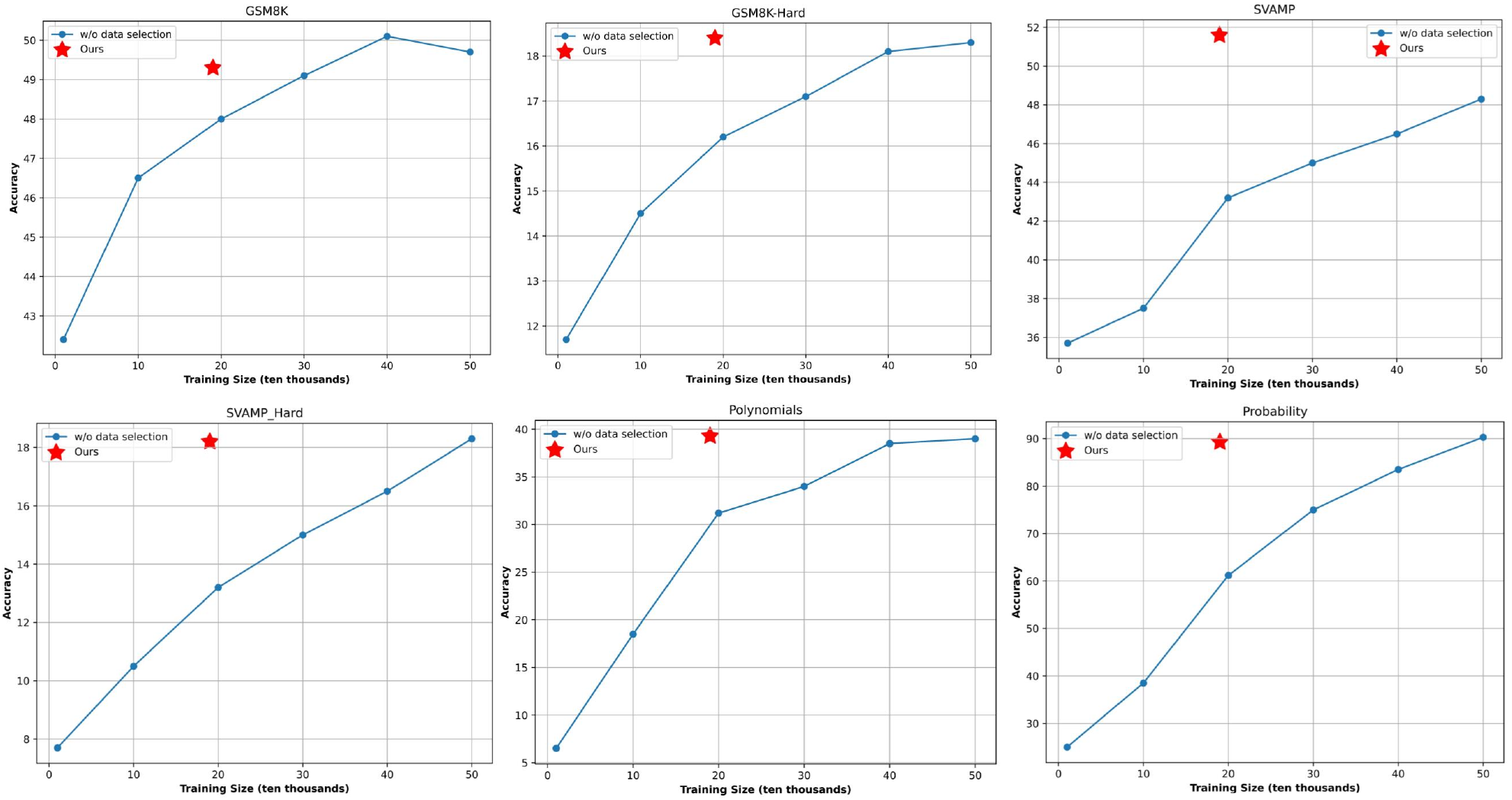}
    \caption{Here, we present the LLaMA 2-7B performances with different size corpus when we don't apply our efficient data selection strategy. Here, we train the model with ControlMathQA and GSM8K.}
    \label{fig:data_selection}
    \vspace{-10pt}
\end{figure*}

\subsubsection{Experimental Settings} 
\paragraph{Training Settings.} The question-answer pairs in ControlMathQA are formatted in Alpaca-format \cite{alpaca}.
Experimentally, we select the LLaMA 2-7B,13B  and Mistral-7B \cite{jiang2023mistral} as backbone models.  We use
a batch size of 128, 512 max token length and train on the ControlMathQA dataset
for 3 epochs using a learning rate of 2e-5 on NVIDIA A100 GPUs. To better illustrate the generalization of our ControlMathQA, we separately  combine it with GSM8K and MetaMath in-domain training datasets to validate its effectiveness in in-domain and out-of-domain results. 

\paragraph{Evaluation Settings.} We evaluate different models in the following datasets: (1) \textbf{SVAMP} \cite{patel-etal-2021-nlp}, an elementary-level math  dataset with 1,000 test examples. (2) \textbf{SVAMP-Hard} \cite{chen2023good}, which replaces numbers in SVAMP questions with values between 100k and 10M. (3) \textbf{GSM8K} \cite{DBLP:journals/corr/abs-2110-14168}, a dataset of 1,391 linguistically diverse grade school MWPs crafted by human writers. (4) \textbf{GSM8K-Hard} \cite{gao2023pal}, an advanced version of GSM8K with larger numerical values to test LLM generalization. (5) The \textbf{Polynomials} and \textbf{Probability} subsets from Mathematics Datasets \cite{saxton2018analysing}, each containing 2,000 samples. (6) \textbf{MATH} \cite{hendrycksmath2021}, which provides competition-level challenges across different mathematical domains.

\subsection{Results}

We evaluate  ControlMathQA to answer the following questions: \textbf{Q1}: Can our dataset promote math generalist LLMs across various domains? \textbf{Q2}: Does our adaptively efficient data selection strategy help model achieve better results with less data?

\begin{figure}
    \centering
    \includegraphics[width=1\linewidth]{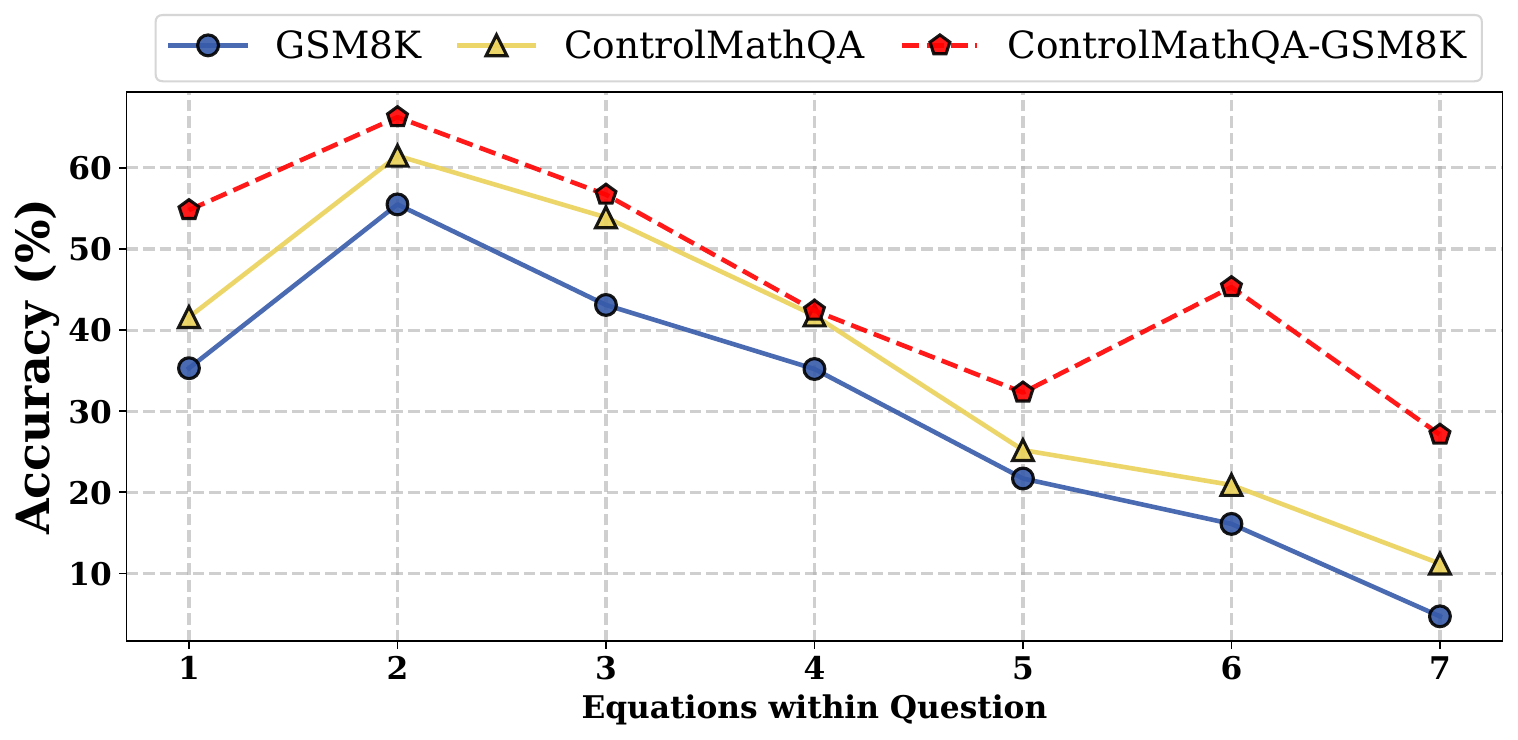}
    \caption{Here, we present the LLaMA 2-7B performances with different training corpus in GSM8K.}
    \label{fig:equation}
    \vspace{-15pt}
\end{figure}

\paragraph{RQ1: ControlMathQA promotes Math Generalist Models.} Table \ref{tab:main_results} presents three backbones' performances with combining ControlMathQA and other in-domain training datasets, separately. Obviously, ours could help different LLMs exhibit exceptional performances across different domains. A notable example is that when training LLaMA 2-7B with ours and GSM8K, the resulting model attains  impressive improvements of 15.8\% and 68.3\% on SVAMP and Probability datasets. Interestingly, though MetaMath could significantly boost the model performances in-domain benchmarks, it downgrades the baseline results in Probability problem solving.
This highlights that overfitting in-domain  helps little or even decreases out-of-domain mathematical capabilities.
These results underscore the effectiveness of our approach in enhancing mathematical generalization. 
% Generated cases of our models are in Appendix \ref{sec:cases}.

\paragraph{RQ2: More data is not always useful.}  In this component, we explore whether our selection strategy can achieve higher training efficiency. Figure \ref{fig:data_selection} shows extensive experiments using GSM8K as the in-domain training corpus without data selection. We can draw the several observations: 1) Scaling up the dataset size generally improves model performance, though the improvement diminishes with larger size. 2) With our data selection strategy, the model achieves much better performances vs. using the same data size without our data selection strategy. 
And mostly ours also achieve better performances when the model trained with 500k data size (except GSM8K).
This highlights that more data is not always helpful and data quality is more important than data quantity.

\input{tables/ablations}

\paragraph{Ablations.} Table \ref{table:ablation2} shows the ablations of different  augmentation problem types in out-of-domain datasets. Obviously, each augmentation contributes to the improvements in different datasets. Multi-step calculation augmented problems lead to better results in SVAMP related datasets. Meanwhile, probability augmentation yields the highest improvements in probability-related tasks.

\subsection{Discussion 1: Tailor ControlMath for specific datasets}
\label{sec:discuss}
Our method can focus on both generalization and adaptation to a targeted dataset. A trivial version of our method involves adaptively generating data based on poor performance on a targeted dataset. Here, we provide an example of how to augment the GSM8K dataset using our method:

First, we train a base SFT model (LLaMA 2-7B) using the GSM8K training set. Then, we test this model on the test set, recording the error rates for problems of varying difficulty levels. The difficulty of a problem is primarily defined by the complexity of the equations involved. Using our ControlMath method, we generate new problems matching the difficulty distribution of the test set problems. We generate more data for problem types with higher error rates, proportional to their contribution to the overall error rate across different difficulty levels. In each iteration, we generate a total of 15,000 new problems and repeat this process three times to iteratively generate the final corpus, termed as \textbf{ControlMathQA-GSM8K. }

Figure \ref{fig:equation} presents the accuracy when the model is trained with  different corpus.  We can find that after such tailored data augmentation, the resulting model significantly improves in all difficulty type questions. Moreover, the overall model performance increases from  42.1\% to 55.3\%, a better result compared with base ControlMathQA (49.3\%). The results indicate that our ControlMath can be regarded as a plug-and-play module to adaptively generate training samples for targeted datasets.

\begin{figure}
\vspace{-10pt}
    \centering
    \includegraphics[width=1\linewidth]{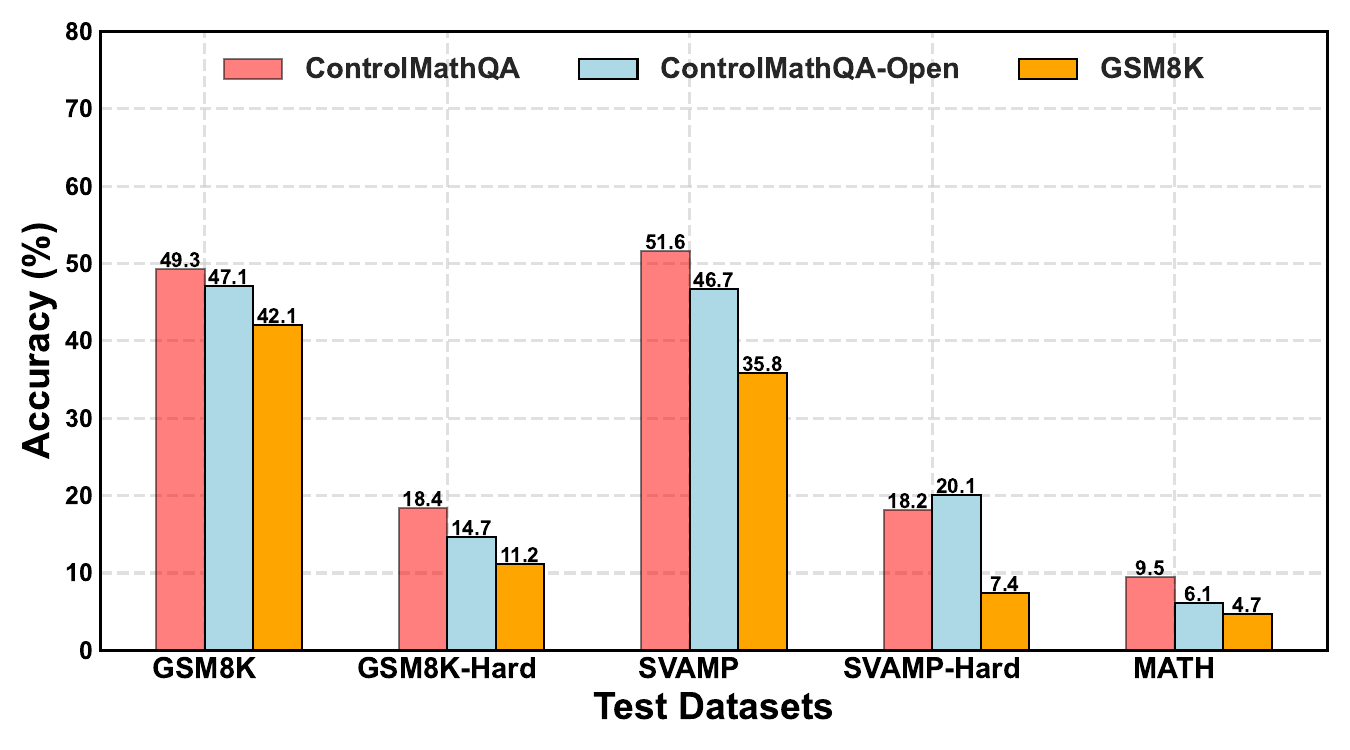}
    \caption{Here, we present the LLaMA 2-7B performances with training  different corpus.}
    \label{fig:efficiency}
    \vspace{-15pt}
\end{figure}

\subsection{Discussion 2: Cost-Effective  ControlMath}

Using GPT-4's API for data generation and validation is costly, resulting in more than 10k dollars. Intuitively, can we use smaller LLMs for data generation to reduce these costs while maintaining high quality? To explore this, we collect related 20k training samples from GPT-4 to train Mistral-7B as the two  Problem-Crafter and Problem-Rewriter Agents.
% To reduce these costs, we intuitively train open-source LLMs to serve as Problem-Crafter and Problem-Rewriter Agents. Specifically, we collect related 20k training samples from GPT-4 to train Mistral-7B as the two agents, which are used to   generate and rewrite mathematical problems.
We also apply rejection sampling to conduct answer augmentation. With the trained Mistral-7B agents, we conduct data augmentation over nearly two months on 8xA100 GPUs, resulting in the collection of approximately 1 million training samples. We term the collected dataset as \textbf{ControlMathQA-Open}.

\begin{figure}
    \centering
    \vspace{-10pt}
    \includegraphics[width=1\linewidth]{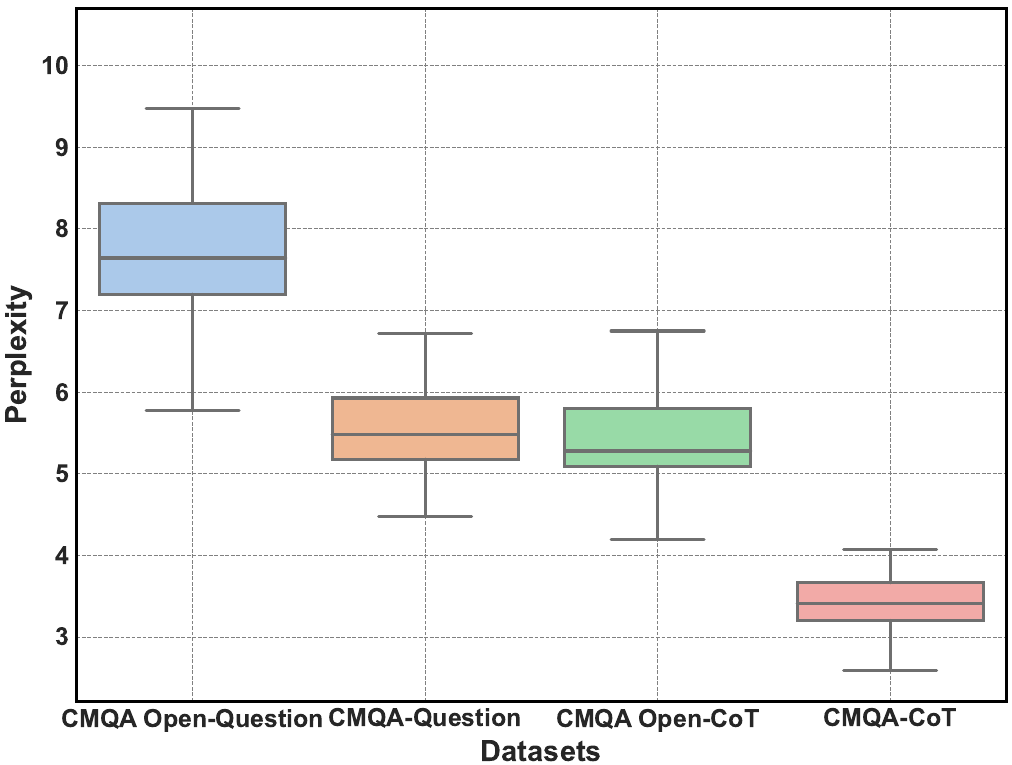}
    \caption{The perplexity of questions and CoT answers in different datasets. CMQA is short for ControlMathQA.}
    \label{fig:ppl}
    \vspace{-1pt}
\end{figure}

Interestingly, as shown in Figure \ref{fig:efficiency}, although ControlMathQA-Open consists of more than five times as much data as the base version, it performs worse across almost five datasets. This suggests that the quality of data generated by Mistral-7B is inferior to that of GPT-4. 
To investigate the reasons behind this, we calculate the perplexity \cite{marion2023more,wang2024making,chen2024oscarsaitheatersurvey} of both datasets using an unfine-tuned Llama 2-7B model.  Figure \ref{fig:ppl} shows that ControlMathQA has a significantly lower perplexity compared to ControlMathQA-Open. This indicates that ControlMathQA is inherently easier to learn from, which might better enhance an LLM's problem-solving abilities \cite{DBLP:journals/corr/abs-2309-12284}.
Moreover, we hypothesize that data augmentation by LLMs inherently resembles knowledge distillation, where a superior teacher model provides more effective problems and answers, thus better facilitating learning. We observe a similar phenomenon during answer augmentation: models trained on answer paths generated by GPT-4 or GPT-3.5 significantly outperform those trained on data from the LLaMA or Mistral series.

% 模型在gsm8k reasoning steps上的表现
% 我们的数据也可能是不准确的

\input{tables/diversity_gain}

\subsection{Discussion 3: Diversity Gains}

we evaluate the diversity of our generated math word problems using the \textit{diversity gain} \cite{bilmes2022submodularitymachinelearningartificial} metric. This metric measures how much a new dataset contributes to the diversity of a base dataset. 
 
 Concretely, given a base dataset $D_{\text{base}}$: Comprises (N) samples, each represented as $(q_i, r_i)$ where $q_i$ is the question, $r_i$ is the CoT answer. And given a new dataset, $D_{\text{new}}$: Comprises (M) samples. Then:
1) For each sample ($x_i$) in $D_{\text{new}}$, find the sample ($x_j$) in $D_{\text{base}}$ that is most similar based on feature extraction.
2) Calculate the squared Euclidean distance in the feature space between ($x_i$) and its nearest ($x_j$) using the formula: ($\min_{x_j \in D_{\text{base}}} (\|f(x_i) - f(x_j)\|^2)$).
3) Average these distances over all samples in $D_{\text{new}}$.

We use gsm8k as base dataset and OpenAI Embedding API text-embedding-ada-002 to extract sentence embeddings. The results are presented in Table \ref{tab:diversity_gain}. ControlMathQA dataset is much more diverse than gsm8k and MetaMath, covering more unseen scenarios.

%% file: tables/main_results.tex
\begin{table*}[t]
    \begin{center}
    \vspace{-15pt}
    \centering
    \small
    % \resizebox{0.8\columnwidth}{!}%
    \begin{tabular}{lcccccccc}
    \toprule
    \multicolumn{9}{c}{\textbf{In-Domain Training Datasets: GSM8K}} \\
\midrule
\multirow{2}{*}{\textbf{Models}}& \multicolumn{2}{|c|}{In-Domain} & \multicolumn{5}{c|}{Out-Of-Domain} & \multirow{2}{*}{\textbf{Avg.}}\\
 % \cmidrule{2-8}
& \multicolumn{1}{|c}{GSM8K} & \multicolumn{1}{c|}{GSM8K-Hard} & \multicolumn{1}{c}{SVAMP} & \multicolumn{1}{c}{SVAMP-Hard} & MATH & Polynomials & \multicolumn{1}{c|}{Prob.} &\\
\midrule 

LLaMA 2-7B& 42.1 & 11.2&35.8 &7.4 &4.7 & 5.5 & 21.0 & 18.2 \\
\textbf{Ours} & 49.3 & 18.4&51.6&18.2&9.5&39.0&89.3 & \textbf{39.3}\\
\midrule
LLaMA 2-13B & 49.7 & 13.5& 50.9 & 14.7 & 6.0&5.5 &22.0 & 23.2 \\
\textbf{Ours} & 57.1 & 16.6& 63.1 & 20.4 &11.2& 41.2 &90.7 & \textbf{42.9} \\

\midrule
Mistral-7B & 51.5 &16.1&37.8&11.3&7.3&4.5&44.9 & 24.8\\
\textbf{Ours} & 59.1& 19.7 &55.7&19.1&12.4&42.1&94.9 &\textbf{43.3}\\
\midrule 
\midrule 
 \multicolumn{9}{c}{\textbf{In-Domain Training Datasets: MetaMath }} \\
\midrule
\multirow{2}{*}{\textbf{Models}}& \multicolumn{3}{|c|}{In-Domain} & \multicolumn{4}{c|}{Out-Of-Domain} & \multirow{2}{*}{\textbf{Avg.}}\\
 % \cmidrule{2-8}
& \multicolumn{1}{|c}{GSM8K} & \multicolumn{1}{c}{GSM8K-Hard} &  \multicolumn{1}{c|}{MATH} &\multicolumn{1}{c}{SVAMP} & \multicolumn{1}{c}{SVAMP-Hard} & Polynomials & \multicolumn{1}{c|}{Prob.} &\\
\midrule 

LLaMA 2-7B&62.9&17.0&19.7&60.8&14.7& 11.9&8.1 & 27.9\\
\textbf{Ours} & 67.0&20.2&21.4&68.9&24.7&42.7&90.3 & \textbf{47.9}\\
\midrule 
LLaMA 2-13B & 66.9 & 18.7& 22.4 & 65.7 & 17.9 &14.0&9.3 &30.7\\
\textbf{Ours} & 70.0 & 22.5& 24.1 & 72.4 & 26.5 &43.4 & 90.1&\textbf{49.9}\\
\midrule 
Mistral-7B & 67.6 & 21.8&22.7 &66.5 & 19.8&20.8&9.1&32.6\\
\textbf{Ours} &71.4&24.5&26.1&71.2& 28.7&50.9&94.7&\textbf{52.5}\\

% LLaMA 2-7B& 1.01&0.50&  0.11& 0.31& 1.71& 0.73 \\

% LLaMA 2-13B& 0.93& 0.66&  0.19& 0.37& 1.76& 0.78\\
% ChatGPT (8k token) &1.30&0.89&0.49&0.29&1.54&0.90 \\
% \midrule
% \textit{Retrieval-based }& \\
% ChatGPT & 1.22& 0.86& 0.37& 0.43& 1.51& 0.88 \\ 

% LLaMA 2-13B&1.73&0.98&0.51&0.24&1.85& 1.06\\
% LLaMA 2-7B& 1.70&0.94&0.54&0.31&1.91& 1.08 \\

% GPT4 &1.91& 0.94& 0.60&0.52&1.69 & 1.13 \\
% \midrule
% \textit{Memory-related}& \\
% MemoryBank-ChatGPT & 1.25& 0.94&0.42&0.45&1.52 &0.92\\
% Resum-ChatGPT & 1.33 &0.97&0.53&0.44&1.49&0.95 \\
% \midrule

% \ModelName-ChatGPT & 1.19&1.07&0.60&0.46&1.62 & 0.99\\ 
% \ModelName-7B& 1.67&1.11&0.60&0.39&1.85 & 1.12 \\
% \ModelName-13B&1.81&1.07&0.70&0.51&1.94 & 1.21\\

% \ModelName-13B DPO&1.79&\textbf{1.20}&\textbf{0.80}&0.46 & \textbf{2.09} & 1.27\\
% \ModelName-GPT4 &\textbf{1.96}&1.14&0.70&\textbf{0.73}&1.85& \textbf{1.28} \\

    \bottomrule
    \end{tabular}
    \end{center}
    \caption{Model performances with different training datasets. Prob. refers to Probability.  Ours denotes that we add our ControlMathQA with specific in-domain training datasets. More comprehensive evaluation of the whole Mathematics and MMLU Datasets can be seen in Appendix, Table \ref{table:mathematics}.}
    \label{tab:main_results}
    \vspace{-15pt}
\end{table*}

%% file: tables/ablations.tex
\begin{table}
% \setlength{\tabcolsep}{1pt}
% \small
\huge
\centering
% \resizebox{\linewidth}{!}
\resizebox{.48\textwidth}{!}{
    % \begin{adjustbox}{raise=-40mm}
    \begin{tabular}{lccccc}
    \toprule[2.2pt] 
    % \makecell{Sythesized \\ Solutions} 
    % \multirow{2}{*}{Data Size} & \multirow{2}{*}{Pseudo-Answer} & \multicolumn{2}{c}{LLaMA} & \multicolumn{2}{c}{CodeLlama} \\ 
    \textbf{Augmentation} & \textbf{SVAMP} & \textbf{SVAMP-H} & \textbf{MATH} & \textbf{Poly.} & \textbf{Prob.} \\ 

    \midrule
    - & 35.8 &7.4 & 4.7&5.5&21.0 \\
    \midrule
    \textbf{w. Multi-step} &47.5&15.9&6.4&9.3&21.2\\
    \textbf{w. Poly.}&39.1&10.3&7.5&37.8&26.0\\
    \textbf{w. Prob.} & 37.5&9.7&7.3&10.1&88.1 \\

    % \specialrule{0em}{-5pt}{0pt} 
    \bottomrule[2.2pt] 
    \end{tabular}}
    % \end{adjustbox}
    \caption{Ablations for different augmentation problems. Here, we train LLaMA 2-7B with our datasets and GSM8K. More results seen in Appendix, Table \ref{table:ablation3}.}
    % 124k_150k2
    \label{table:ablation2}
    \vspace{-15pt}
    % \vspace{-12mm}
\end{table}

%% file: tables/diversity_gain.tex
\begin{table}[]
    \centering
    \begin{tabular}{l|ccc}
    \toprule
       Dataset  & gsm8k & MetaMath & Ours \\
       \midrule
     Diversity Gain    & -&0.09&0.41\\
     \bottomrule
    \end{tabular}
    \caption{The diversity gain between MetaMath and ours}
    \label{tab:diversity_gain}
\end{table}

%% file: Section/5Conclusion.tex
\section{Conclusion}

In this paper, we propose a new data augmentation approach for mathematical reasoning, called ControlMath. The core of this approach lies into two lines: 1) It first utilizes an equation generation module to control the distributions of the generated equations. Then a Probelm-Crafter Agent generates the math word problems based on these equations. 2) It introduces a Problem-Rewriter Agent to help ensure the effectiveness of the generated samples, filtering out redundant ones. Our approach could generate diverse problems and focus on quality over quantity to enable stronger math generalist models with smaller data. The resulting dataset, ControlMathQA, could help LLMs obtain improvements in both in-domain and out-of-domain datasets. Future works include expanding the size of ControlMathQA and exploring other strategies to select more efficient data.

%% file: tables/prompts.tex
% \begin{table*}[!t]\footnotesize
% \centering
% \small

% \begin{tabular}{p{0.95\linewidth}}
% % \begin{tabular}{lc}
% % \multirow{28}{*}{\textbf{Persona Prompts}} & 
% \toprule
% \multicolumn{1}{c}{Prompt that is used for obtaining memory-grounded responses for GPT4-Turbo.} \\
% \midrule
% You are a language expert. Your task is to rewrite the given question.  \\
% The following are several examples: \\
% \\

% \bottomrule
% \caption{Prompts that are used for obtaining DPO samples in Task 3. Only English translation is provided for 
% easing reading.} 
% \label{table:dpoprompt}
% \end{tabular}

% \vspace{-5mm}
% \end{table*}
% \tcbset{
%   enhanced jigsaw, 
%   colback=red!5!white, 
%   colframe=red!75!black, 
%   boxrule=1pt,
%   breakable % 允许跨栏
% }

% \lipsum[1] % 生成示例文本
% \begin{tcolorbox}[
% breakable,
% colback=white!10!white,
% colframe=black!75!black,
% title=Fine-tuning system prompt]

\onecolumn
\begin{exmp}{Prompts for Problem-Rewriter Agents}{exmp:agent}
You are a language expert. Your task is to rewrite the given question to make it more diverse by rephrasing or changing the topic.  \\
The following are several examples: \\
\\
\textbf{Question}: Sam bought a dozen boxes, each with 30 highlighter pens inside, for \$10 each box. He rearranged
five of these boxes into packages of six highlighters each and sold them for \$3 per package. He sold the
rest of the highlighters separately at the rate of three pens for \$2. How much profit did he make in total, in
dollars? \\
\textbf{New Question}: Sam purchased 12 boxes, each containing 30 highlighter pens, at \$10 per
box. He repackaged five of these boxes into sets of six highlighters and sold them for \$3 per set. He sold
the remaining highlighters individually at a rate of three pens for \$2. What is the total profit he made in dollars? \\
\\
\textbf{Question}: A group of people went to
an amusement park. 37 of them were
children and 47 were adults. They
split the total number of people into
two groups and added 18 more people
to one of the groups. This group was
then divided equally among six rides.
How many people rode on each ride? \\
\textbf{New question}: A school organized a field trip to a local museum. Initially, 37 students and 47 teachers attended. The entire group was divided into two, and then 18 volunteers joined one of the groups. This larger group was evenly distributed across six exhibit tours. How many people were in each tour group?
\\ 
\\
\textbf{Question}:  Express -41*c - 16*c**2 + 18*c + 25*c in the form q*c**2 + p*c + u and give p \\
\textbf{New question}: 
Rewrite the polynomials -41*x - 16*x**2 + 18*x + 25*x in the form a*x**2 + b*x + c and determine the coefficient b.
\\
\\
\textbf{Question}: What is probability of picking 1 k, 1 h, and 1 c when three letters picked without replace-
ment from \{c: 1, y: 1, e: 1, n: 1, k: 1, h: 2\}? \\
\textbf{New question}: What are the odds of drawing one k, one h, and one c from a collection that includes one each of c, y, e, n, k, and two h, if three letters are chosen sequentially without putting any back? \\
\\

Question: \{\texttt{Question}\}\\

You should keep in mind that you can not change the numbers and operators in the question.

% \caption{Query augmentation prompt.}
% \label{tab:query-aug prompt}
\end{exmp}

\begin{exmp}{Prompts for GPT-4 Evaluation}{exmp:gpt4}
You are a math expert. Given a question, a reference solution, and an equation, assess whether all components are logically consistent and correctly matched.   Ensure that the question, solution, and equation are congruent. This means that:
1) The solution should correctly answer the question.
2) The equation should be applicable and necessary for deriving the solution. 3) 
All mathematical operations and logic used in the solution should be supported by the equation. \\

Question: \{\texttt{Question}\} \\
Reference Solution: \{\texttt{Solution}\} \\
Equation: \{\texttt{Equation}\} \\

If the question, reference solution and equation all align and math correctly, please output \textbf{True}. Otherwise, output \textbf{False}.

% \caption{Query augmentation prompt.}
% \label{tab:query-aug prompt}
\end{exmp}

\begin{exmp}{Prompts for Fine-tuning}{exmp:finetuning}
Below is an instruction that describes a task. Write a response that appropriately completes the request.\#\#\# Instruction:
 \texttt{\{Query\}} \#\#\# Response:
% \caption{Query augmentation prompt.}
% \label{tab:query-aug prompt}
\end{exmp}

\begin{figure*}
    \centering
    \includegraphics[width=1\linewidth]{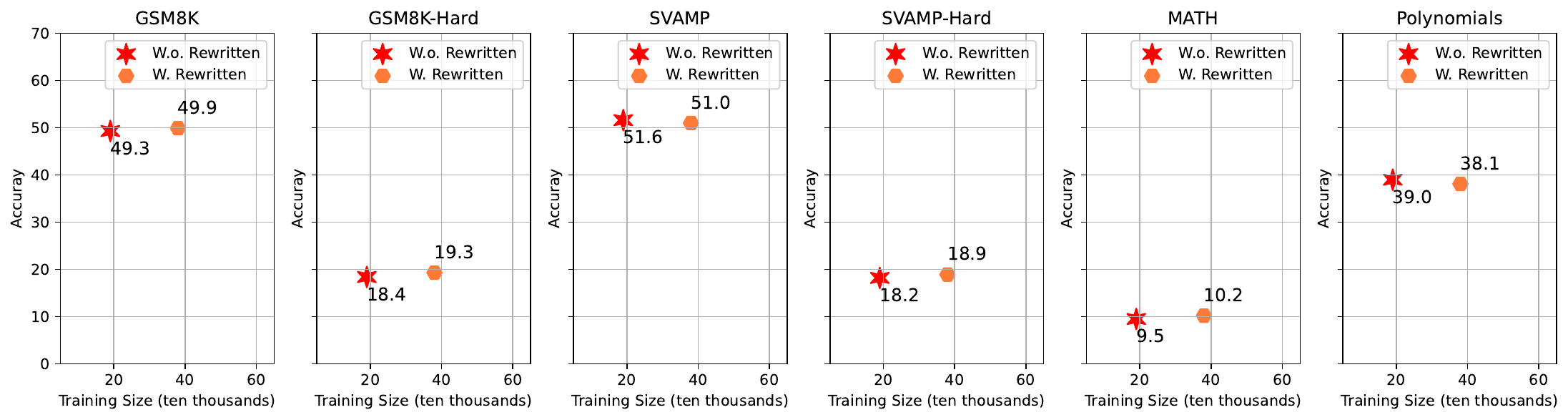}
    \caption{Here, we present the LLaMA 2-7B performances with including rewritten problems vs. without rewritten problems (Ours). Here, we train the model with ControlMathQA and GSM8K.}
    \label{fig:rewritten}
    \vspace{-10pt}
\end{figure*}

\input{tables/confusion_matrix}

\input{tables/abl_all}

\twocolumn

% what are you talking

%% file: tables/confusion_matrix.tex
\begin{table*}[h]
\centering
\small
\begin{tabular}{c|c|c|c|c}

\toprule
Problem Typs&\multicolumn{4}{c}{Original Problem } \\
\midrule
\multirow{5}{*}{Rewritten Problems} & Percentage (\%) &Wrong & Correct &Total \\
\cmidrule{2-5}
% \midrule
&Wrong &  33.1 & 13.4 &46.5\\
\cmidrule{2-5}
&Correct & 6.5 & 47 &53.5\\
\cmidrule{2-5}
& Total &39.6&60.4&100 \\
\bottomrule
\end{tabular}
\caption{Confusion matrix of the proportions between rewritten and original problems in total three iterative processes.
}
\vspace{-15pt}
\label{pilot}

\end{table*}

% \multirow{4}{*}{zero-shot} 

%% file: tables/abl_all.tex
\begin{table*}
% \setlength{\tabcolsep}{1pt}
% \small
% \huge
\centering
% \resizebox{\linewidth}{!}
\resizebox{.99\textwidth}{!}{
    % \begin{adjustbox}{raise=-40mm}
    \begin{tabular}{lccccccc}
    \toprule[2.2pt] 
    % \makecell{Sythesized \\ Solutions} 
    % \multirow{2}{*}{Data Size} & \multirow{2}{*}{Pseudo-Answer} & \multicolumn{2}{c}{LLaMA} & \multicolumn{2}{c}{CodeLlama} \\ 
    \textbf{Augmentation} & \textbf{GSM8K}& \textbf{GSM8K-Hard}&\textbf{SVAMP} & \textbf{SVAMP-H} & \textbf{MATH} & \textbf{Poly.} & \textbf{Prob.} \\ 

    \midrule
    - & 42.1&11.2&35.8 &7.4 & 4.7&5.5&21.0 \\
    \midrule
    \textbf{w. Multi-step} &47.1&16.3&47.5&15.9&6.4&9.3&21.2\\
    \textbf{w. Poly.}&45.1&14.3&39.1&10.3&7.5&37.8&26.0\\
    \textbf{w. Prob.} & 44.3&13.7&37.5&9.7&7.3&10.1&88.1 \\

    % \specialrule{0em}{-5pt}{0pt} 
    \bottomrule[2.2pt] 
    \end{tabular}}
    % \end{adjustbox}
    \caption{Ablations for different augmentation problems. Here, we train LLaMA 2-7B with our datasets and GSM8K.}
    % 124k_150k2
    \label{table:ablation3}
    % \vspace{-12mm}
\end{table*}

\begin{table*}
% \setlength{\tabcolsep}{1pt}
% \small
% \huge
\centering
% \resizebox{\linewidth}{!}
\resizebox{.99\textwidth}{!}{
    % \begin{adjustbox}{raise=-40mm}
    \begin{tabular}{lccccccccc}
    \toprule[2.2pt] 
    % \makecell{Sythesized \\ Solutions} 
    % \multirow{2}{*}{Data Size} & \multirow{2}{*}{Pseudo-Answer} & \multicolumn{2}{c}{LLaMA} & \multicolumn{2}{c}{CodeLlama} \\
    \textbf{Methods} &	\textbf{Algebra}	&\textbf{Arithmetic}	& \textbf{Calculus}	& \textbf{Numbers}	&\textbf{Comparison}&	\textbf{Polynomials}	&\textbf{Probability}&	\textbf{MMLU-Stem}&	\textbf{Avg.}\\

    \textbf{GSM8K} &	9.5	&46.5	&3.3	&35.4&	70.1	&5.5	&21.0&	29.7	&27.6 \\
MetaMath&	13.2&	61.2	&8.8	&60.1	&77.5	&11.9	&8.1&	42.6	&35.5 \\
\textbf{Ours}&	\textbf{74.5}&	\textbf{70.6}	&\textbf{28.9}	&\textbf{87.3}	&\textbf{91.4}	&\textbf{89.3}	&\textbf{39.3}	&\textbf{46.3}	&\textbf{65.9} \\

    % \specialrule{0em}{-5pt}{0pt} 
    \bottomrule[2.2pt] 
    \end{tabular}}
    % \end{adjustbox}
    \caption{Evaluation results in Mathematics Dataset and MMLU-Stem. Here, the models are trained on LLaMA2-7B.}
    % 124k_150k2
    \label{table:mathematics}
    % \vspace{-12mm}
\end{table*}

%% file: tables/cases.tex
\onecolumn
\section{Cases}
\label{sec:cases}
\begin{case}{Several cases of our models solving Probability problems}{case:finetuning}
\textbf{Question}: Three letters picked without replacement from {y: 3, a: 1, p: 3, v: 2, c: 2, t: 2}. Give prob of sequence cca. \\

\textbf{Answer}: To find the probability of picking the sequence ``cca'' without replacement from the set $\{y: 3, a: 1, p: 3, v: 2, c: 2, t: 2\}$, we calculate the probability step by step:
The total number of letters is $3 + 1 + 3 + 2 + 2 + 2 = 13$.
The probability of picking ``c'' first is $\frac{2}{13}$ since there are 2 ``c''s out of 13 letters.
After picking the first ``c'', there are now 12 letters left, and the probability of picking another ``c'' is $\frac{1}{12}$.
After picking the second ``c'', there are now 11 letters left, and the probability of picking ``a'' is $\frac{1}{11}$.
The probability of the sequence ``cca'' is the product of these probabilities:
$$\frac{2}{13} \times \frac{1}{12} \times \frac{1}{11} = \frac{1}{858}$$
Therefore, the probability of picking the sequence ``cca'' without replacement is $\frac{1}{858}$.
Therefore, the answer is 1/858.
\#\#\# 1/858. \\

\textbf{Question}: Two letters picked without replacement from wdwwwwwrdwrrww. Give prob of sequence dr. \\

\textbf{Answer}: To find the probability of picking the sequence ``dr'' without replacement from the string ``wdwwwwwrdwrrww'', we calculate as follows:
The total number of letters is $14$.
There are $2$ ``d''s and $3$ ``r''s in the string.
The probability of picking ``d'' first is $\frac{2}{14}$ since there are $2$ ``d''s out of $14$ letters.
After picking ``d'', there are now $13$ letters left, and the probability of then picking ``r'' is $\frac{3}{13}$, as there are $3$ ``r''s remaining.
The probability of the sequence ``dr'' is the product of these probabilities:
$$\frac{2}{14} \times \frac{3}{13} = \frac{3}{91}$$
Therefore, the probability of picking the sequence ``dr'' without replacement is $\frac{3}{91}$.
Therefore, the answer is 3/91.
\#\#\# 3/91.
\end{case}

% \section{Cases}
\begin{case}{Several cases of our models solving Polynomial problems}{case:finetuning}
\textbf{Question}: Express -37 - f**3 + 70 + 26 as b + w*f**2 + x*f + y*f**3 and give b.
\\

\textbf{Answer}:  Given equation: $-37 - f^3 + 70 + 26$
Simplify the constants: $-37 + 70 + 26 = 59$
Therefore, the equation can be rewritten as: $59 - f^3$
In the form $b + wf^2 + xf + yf^3$, we have:
\\
$b = 59$\\
$w = 0$ (since there is no $f^2$ term)\\
$x = 0$ (since there is no $f$ term)\\
$y = -1$ (coefficient of $f^3$)\\
Therefore, the answer is 59.
\#\#\# 59. \\

\textbf{Question}: Frankie, a young scientist, is experimenting with temperature effects on a certain chemical compound. She measures the change in energy, denoted as $E$, when a sample is cooled. The energy change depends on the temperature in Fahrenheit, $f$, according to the equation:
\[
E = -37 - f^3 + 70 + 26
\]

Frankie records a temperature of $-2$ Fahrenheit during one of her experiments. Calculate the energy change for this temperature using the simplified equation.
 \\

\textbf{Answer}: Substituting $f = -2$ into the simplified equation:
    \[
    E = -(-2)^3 + 59 = -(-8) + 59 = 8 + 59 = 67
    \]
    The energy change at $f = -2$ Fahrenheit is $67$ units. Therefore the answer is 67 \#\#\#67.
\end{case}

\begin{case}{Several cases of our models solving GSM8K/GSM8K-Hard problems}{case:finetuning}
\textbf{Question}: Susan made 100 cookies for Christmas and was going to equally divide them between her 6 nephews. Before Susan could package them, her husband snuck 4 cookies for himself. How many cookies will each of Susan’s nephews get?
\\

\textbf{Answer}:  Susan made 100 cookies in total.Her husband snuck 4 cookies for himself, so there are <<100 - 4 = 96>> cookies left.Susan wants to divide these cookies equally between her 6 nephews.To find out how many cookies each nephew will get, we divide the total number of cookies by the number of nephews: <<96 / 6 = 16>>. Each of Susan's nephews will get 16 cookies.
Therefore, the answer is 16.
\#\#\# 16. \\

\textbf{Question}: Siobhan has 2 fewer jewels than Aaron. Aaron has 5 more jewels than half of Raymond's jewels. If Raymond has 973054 jewels, how many jewels does Siobhan have? \\

\textbf{Answer}: We know that Raymond has 973054 jewels.Half of Raymond's jewels is <<973054/2 = 486527>> jewels.Aaron has 5 more jewels than half of Raymond's jewels, so Aaron has <<486527 + 5 = 486532>> jewels.Siobhan has 2 fewer jewels than Aaron, so Siobhan has <<486532 - 2 = 486530>> jewels.Therefore, Siobhan has 486530 jewels.  Therefore, the answer is 486530. \#\#\#486530.
\end{case}